\documentclass[letterpaper, 10 pt, conference]{ieeeconf}  %

\usepackage{amssymb}
\usepackage[per=frac,binary-units=true]{siunitx}
\usepackage[pdftex]{graphicx}

\newcounter{todo}

\newcommand{\etal}{et~al.}
\newcommand{\ie}{i.e.,\ }
\newcommand{\eg}{e.g.,\ }

\usepackage{booktabs}

\usepackage{amsmath} %
\usepackage{amssymb}  %

\usepackage{subcaption} %

\graphicspath{{figures/}}

\newlength{\figurewidth}
\newlength{\figureheight}
\usepackage{booktabs} %
\usepackage{hyperref}
\usepackage[capitalize,noabbrev]{cleveref}
\crefname{algocf}{Algorithm}{Algorithms}
\crefname{table}{Table}{Tables}
\crefname{chapter}{Chapter}{Chapters}
\crefname{equation}{Equation}{Equations}
\crefname{section}{Section}{Sections}

\usepackage{tikz}
\usetikzlibrary{arrows}
\usetikzlibrary{positioning,calc}
\usetikzlibrary{decorations.pathreplacing}
\usetikzlibrary{decorations.markings}
\usetikzlibrary{fit}
\usetikzlibrary{shapes.callouts}
\usepackage{pgfplots}

\tikzset{
  big arrow/.style={
    decoration={markings,mark=at position 1 with {\arrow[scale=1.7]{latex}}},
    postaction={decorate},
    shorten >=0.4pt}}
\tikzset{
  big 2arrow/.style={
    decoration={markings,mark=at position 0 with {\arrow[scale=-1.7]{latex}},mark=at position 1 with {\arrow[scale=1.7]{latex}}},
    postaction={decorate},
    shorten >=0.4pt}}

\IEEEoverridecommandlockouts                              %

\usepackage{graphics} %

\title{\LARGE \bf
Efficient Continuous-time SLAM for 3D Lidar-based Online Mapping
}

\author{David Droeschel and Sven Behnke%
\thanks{All authors are with the Autonomous Intelligent Systems Group, Computer Science Institute VI,
        University of Bonn, 53115 Bonn, Germany
        {{\tt \{droeschel,~behnke\}@ais.uni-bonn.de}}}%
        }

\begin{document}

\maketitle
\thispagestyle{empty}
\pagestyle{empty}

\begin{tikzpicture}[remember picture,overlay]
	\node[anchor=north west,align=left,font=\sffamily,yshift=-0.2cm] at (current page.north west) {%
		  In: Proceedings of the International Conference on Robotics and Automation (ICRA) 2018
		  };
		  \node[anchor=north east, align=right,font=\sffamily,yshift=-0.2cm] at (current page.north east) {%
			    DOI: \href{https://doi.org/10.1109/ICRA.2018.8461000}{10.1109/ICRA.2018.8461000}
			    };
\end{tikzpicture}%

\begin{abstract}

Modern 3D laser-range scanners have a high data rate, making online simultaneous localization and mapping (SLAM) computationally challenging. 
Recursive state estimation techniques are efficient but commit to a state estimate immediately after a new scan is made, which may lead to misalignments of measurements. 
We present a 3D SLAM approach that allows for refining alignments during online mapping.
Our method is based on efficient local mapping and a hierarchical optimization back-end.
Measurements of a 3D laser scanner are aggregated in local multiresolution maps by means of surfel-based registration. The local maps 
are used in a multi-level graph for allocentric mapping and localization. 
In order to incorporate corrections when refining the alignment, the individual 3D scans in the local map are modeled as a sub-graph and graph optimization is performed to account for drift and misalignments in the local maps.
Furthermore, in each sub-graph, a continuous-time representation of the sensor trajectory allows to correct measurements between scan poses.
We evaluate our approach in multiple experiments by showing qualitative results. 
Furthermore, we quantify the map quality by an entropy-based measure.

\end{abstract}

\section{Introduction}

Laser-based mapping and localization has been widely studied in the robotics community and applied to many robotic platforms~\cite{Nuchter05,magnusson2007_3dndt,Kohlbrecher2011,Cole2006}. The variety of approaches that exists either focus on efficiency, for example when used for autonomous navigation, or on accuracy when building high-fidelity maps offline. Often, limited resources---such as computing power on a micro aerial vehicle---necessitate a trade-off between the two. A popular approach to tackle this trade-off is to leverage other sensor modalities to simplify the problem. For example, visual odometry from cameras and inertial measurement units~(IMU) are used, to estimate the motion of the laser-range sensor over short time periods. The motion estimate is used as a prior when aligning consecutive laser scans, allowing for fast and relatively accurate mapping. 

Often inaccuracies remain, for example caused by wrong data associations in visual odometry. These inaccuracies lead to misalignments and degeneration in the map and require costly reprocessing of the sensor data. To this end, graph-based optimization is popular to minimize accumulated errors~\cite{kuemmerle2011_g2o,Frese1416971,olson2006fast}. However, depending on the granularity of the modeled graph, optimization is computationally demanding for large maps.

Another difficulty in laser-based SLAM is the sparseness and distribution of measurements in laser scans. As a result, pairwise registration of laser scans quickly accumulates errors. 
Registering laser scans to a map, built by aggregating previous measurements, often minimizes accumulated error. 

However, errors remain, \eg due to missing information. For example, incrementally mapping the environment necessitates bootstrapping from sparse sensor data at the beginning---resulting in relatively poor registration accuracy, compared to aligning with a dense and accurate map.

In our previous work~\cite{Droeschel2017104}, we showed that local multiresolution in combination with a surfel-based registration method allows for efficient and robust mapping of sparse laser scans. 
The key data structure in our previous work is a robot-centric multiresolution grid map to recursively aggregate laser measurements from consecutive 3D scans, yielding local mapping with constant time and memory requirements. Furthermore, modeling a graph of local multiresolution maps allows for allocentric mapping of large environments~\cite{Droeschel:JFR2015}. 
While being efficient, the approach did not allow reassessing previously aggregated measurements in case of registration errors and poor or missing motion estimates from visual odometry. 

\begin{figure}[t]
\centering 
\vspace{1em}
{\includegraphics[width=0.99\linewidth]{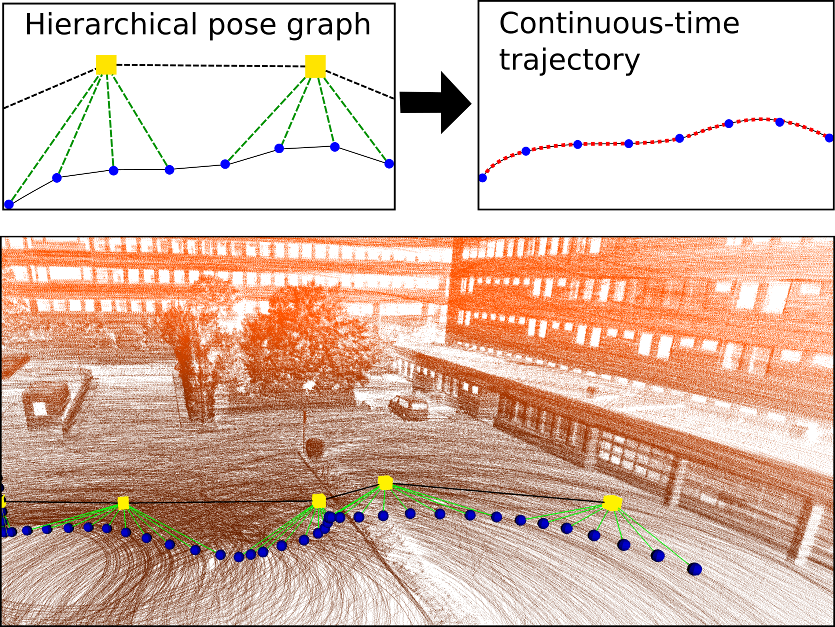}}
\caption{
We propose a hierarchical continuous-time SLAM method, allowing for online map refinement.
It generates highly accurate maps of the environment from laser measurements. Yellow squares: coarse nodes; Blue circles: fine nodes; Red dots: continuous-time trajectory.} 
\label{fig:teaser}
\end{figure}

 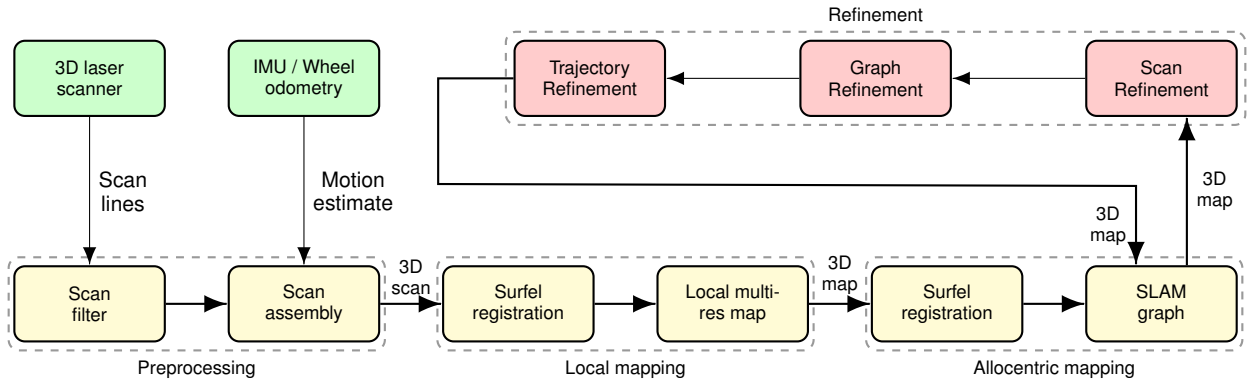
\begin{figure*}
\begin{center}
  \begin{tikzpicture}
    [
    font=\sffamily\scriptsize,
    n/.style={draw, thick, rounded corners,fill=yellow!20,minimum height=1cm, minimum width=2cm},
    sensor/.style={n,fill=green!20},
    op/.style={n,fill=red!20},
    module/.style={dashed,draw=black!40,thick,rounded corners,inner sep=0.2cm,transform shape,inner sep=3pt},
    every node/.style={
      align=center
    },
    xscale=0.95
  ]

  \node[n] (scanfilter) at (0.0, 0.0) {Scan\\filter};
  \node[n] (assembler) at (3.0, 0.0) {Scan\\assembly};
  \node[n] (registration) at (6.0, 0.0) {Surfel\\registration};
  \node[n] (map) at (9.0, 0.0) {Local multi-\\res map};
  \node[n] (registration2) at (12.0, 0.0) {Surfel\\registration};
  \node[n] (gmap) at (15.0,0.0) {SLAM\\graph};

  \node[module,fit={($(scanfilter)-(33pt,0)$) (assembler) ($(assembler)+(33pt,0)$)}] (preprocessing) {};
  \node[below = 0.0cm of preprocessing] {Preprocessing};

  \node[module,fit={ ($(registration)-(33pt,0)$) (registration) (map) ($(map)+(33pt,0)$)}] (local_mapping) {};
  \node[below = 0.0cm of local_mapping] {Local mapping};

 \node[module,fit={($(registration2)-(33pt,0)$) (registration2) (gmap) ($(gmap)+(33pt,0)$) }] (gmapping) {};
 \node[below = 0.0cm of gmapping] {Allocentric mapping};


  \draw[big arrow,thick] (scanfilter) -- (assembler);
  \draw[big arrow,thick] (assembler) -- (registration) node [midway, above] {3D\\scan};
  \draw[big arrow,thick] (registration) -- (map);
  \draw[big arrow,thick] (map) -- (registration2) node[midway,above] {3D\\map};
  \draw[big arrow,thick] (registration2) -- (gmap) node[midway,above] {};

  \node[sensor] (laser) at (0.0, 3.0) {3D laser\\scanner};
  \draw[big arrow] (laser) -- (scanfilter) node [midway, right,font=\footnotesize\sffamily] {Scan\\lines};

  \node[sensor] (imu) at (3.0, 3.0) {IMU / Wheel\\odometry};

  \draw[big arrow] (imu) -- (assembler) node [midway, right,font=\footnotesize\sffamily] {Motion\\estimate};





  \node[op] (graph_refinement) at (11.0, 3.0) {Graph\\Refinement}; 
  \node[op] (scan_refinement) at (15.0, 3.0) {Scan\\Refinement}; 
  \node[op] (scanline_refinement) at (7.0, 3.0) {Trajectory\\Refinement}; 
  \draw[big arrow] (scan_refinement) -- (graph_refinement) node [midway, below,font=\footnotesize\sffamily] {};
  \draw[big arrow] (graph_refinement) -- (scanline_refinement) node [midway, above,font=\footnotesize\sffamily] {};

 \node[module,fit={($(scanline_refinement)-(38pt,0)$) (graph_refinement) ($(scan_refinement)+(38pt,0)$)  }] (refinement) {};
 \node[above = 0.0cm of refinement] {Refinement};

  \draw[big arrow,thick] ($(gmap.north west)!2/3!(gmap.north east)$) --($(scan_refinement.south west)!2/3!(scan_refinement.south east)$) node[midway,right] {3D\\map};
  \draw[big arrow,thick] (scanline_refinement.west)--($(scanline_refinement.west)-(30pt,0)$) --($(scanline_refinement.west)-(30pt,40pt)$)--($(gmap.north west)!1/3!(gmap.north east)+(0,30pt)$)--($(gmap.north west)!1/3!(gmap.north east)$) [out=150, in=20] [out=150, in=20] [bend left] node[midway,left] {3D\\map};
\end{tikzpicture}
\end{center}
\caption{Schematic illustration of our mapping system. Laser measurements are preprocessed and assembled to 3D point clouds. The resulting 3D point cloud is used to estimate the transformation between the current scan and map. Registered scans are stored in a local multiresolution map. Local multiresolution maps from different view poses are registered against each other in a SLAM graph. During mapping, parts of the graph are refined and misaligned 3D scans are corrected. }
\label{fig:mapping_architecture}
\end{figure*} 

In this paper, we extend our previous approach, allowing for reassessing the registration of previously added 3D scans. By modeling individual 3D scans of a local map as a sub-graph, we build a hierarchical graph structure, enabling refinement of the map in case misaligned measurements when more information is available. Furthermore, the approach preserves efficient local and allocentric mapping, as with our previous method.
In summary, the contribution of our work is a novel combination of a hierarchical graph structure---allowing for scalability and efficiency---with local multiresolution maps to overcome alignment problems due to sparsity in laser measurements, and a continuous-time trajectory representation.

\begin{figure*}
\centering
\fbox
{
\includegraphics[width=0.95\textwidth, trim=0cm 0cm 0cm 0.0em, clip]{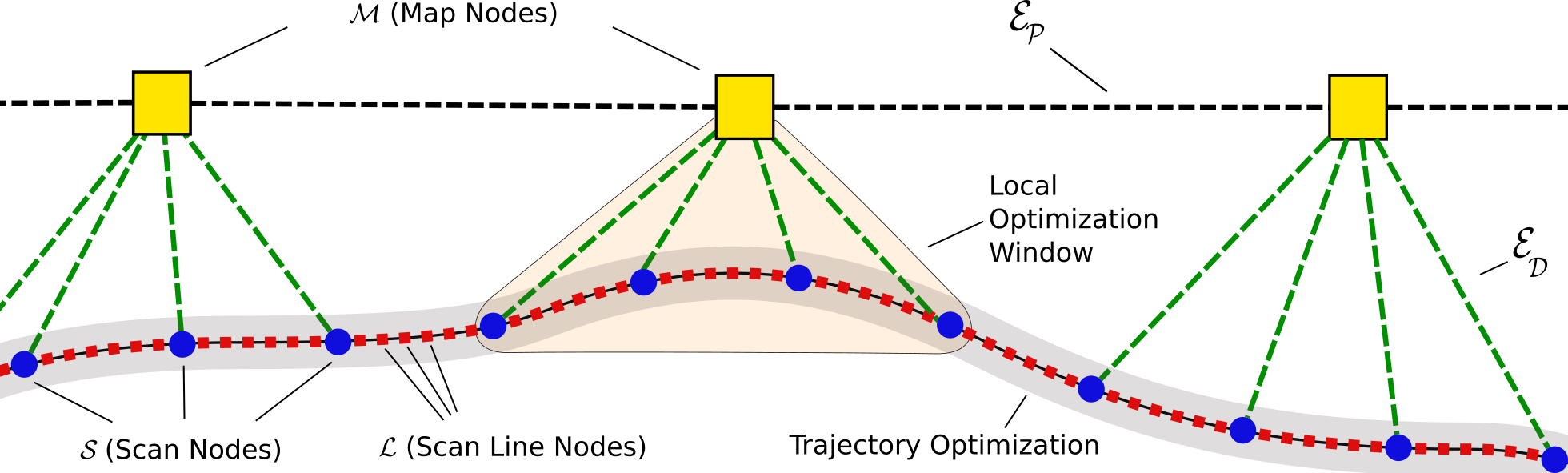}
}\\
\caption{Hierarchical graph representation of the optimization problem at hand: The vertex sets $\mathcal{M}, \mathcal{S}$, and $\mathcal{L}$ correspond to the estimation variables, \ie the poses of the local multiresolution maps ($\mathcal{M}$), the 3D scans ($\mathcal{S}$), and   scan lines ($\mathcal{L}$) of the Velodyne~VLP-16. The edge sets $\mathcal{E}_\mathcal{P}$ and $\mathcal{E}_\mathcal{D}$ represent constraints from registration: $\mathcal{E}_\mathcal{P}$ from aligning two local maps to each other and $\mathcal{E}_\mathcal{D}$ from aligning a 3D scan to a surfel map. From $\mathcal{S}$ a continuous-time representation of the trajectory is estimated by a cubic B-spline, allowing to interpolate the pose for each measurement of the 3D scan ($\mathcal{L}$). }
\label{fig:graph}
\end{figure*}

\section{Related Work}

Mapping with 3D laser scanners has been investigated by many groups~\cite{Nuchter05,magnusson2007_3dndt,Kohlbrecher2011,Cole2006}.
While many methods assume the robot to stand still during 3D scan acquisition, some approaches also integrate scan lines of a continuously rotating laser scanner into 3D maps while the robot is moving~\cite{bosse2009_continuous_scanmatching,elseberg2012_semirigid_slam,stoyanov2009_mlpca,maddern2012_lostintranslation,anderson2013_continuous_slam}.
The mentioned approaches allow creating accurate maps of the environment under certain conditions, but do not allow an efficient assessment and refinement of the map.

Measurements from laser scanners are usually subject to rolling shutter artifacts when the sensor is moving during acquisition. 
These artifacts are expressed by a deformation of the scan and, when treating laser scans as rigid bodies, these artifacts degrade the map quality and introduce errors when estimating the sensor pose. 
A common approach to address this problem is to model a deformation in the objective function of the registration approach. 
Non-rigid registration of 3D laser scans has been addressed by several groups~\cite{brown2007global, elseberg2010non,ruhnke2012highly,zlot2014efficient,Anderson2015BatchNC}.

Ruhnke~\etal~\cite{ruhnke2012highly} jointly optimize sensor poses and measurements. They extract surface elements from range scans, and seek for close-by surfels from different scans. This data association contributes to the error term of the optimization problem but also results in a relatively high state space. Thus, their approach can build highly accurate 3D maps but does not allow for online processing.

Furthermore, rolling shutter effects can be addressed by modeling the sensor trajectory as a continuous function over time, instead of a discrete set of poses.
Continuous-time representations show great advantages when multiple sensors with different temporal behavior are calibrated~\cite{furgale2013unified} or fused~\cite{mueggler2017continuous}, but also to compensate for rolling shutter effects, \eg for data from a RGB-D camera~\cite{kerl2015dense}. Continuous-time trajectory representations have been used for laser-based mapping in different works~\cite{alismail2014continuous,patron2015spline,KaulROB21614}. 
While most of the continuous-time approaches use a spline to represent the trajectory, Anderson \etal~\cite{Anderson2015BatchNC} employ Sparse Gaussian Process Regression. 

Kaul~\etal~\cite{zlot2014efficient,KaulROB21614} present a continuous-time mapping approach using non-rigid registration and global optimization to estimate the sensor trajectory from a spinning laser scanner and an industrial-grade IMU. The trajectory is modeled as a continuous function and a spline is used to interpolate between the sensor poses. 

Recently, Hess~\etal~presented \textit{Google's Cartographer}~\cite{7487258}. 
By aggregating laser scans in local 2D grid maps and an efficient branch-and-bound approach for loop closure optimization. 
Results of Google's Cartographer have been improved by N\"uchter~\etal~\cite{nuchter2017improving}. 
Their method refines the resulting trajectory by a continuous-time mapping approach, based on their previous work~\cite{elseberg2013algorithmic}.

Grisetti~\etal~present a hierarchical graph-based SLAM approach~\cite{5509407}. 
Similar to our approach, a hierarchical pose graph represents the environment on different levels, allowing for simplifying the problem and optimizing parts of it independently.

Following Grisetti~\etal~\cite{5509407}, we model our problem as hierarchical graph, allowing us to optimize simplified parts of the problem independently. 
Compared to their approach we aggregate scans in local sub-maps to overcome sparsity in the laser scans. 
Furthermore, we augment the local sub-graphs with a continuous-time representation of the trajectory, allowing to address the mentioned rolling shutter effects.

\section{System Overview}

Our system aggregates measurements from a laser scanner in a robot-centric local multiresolution grid map---having a high resolution in the close proximity to the sensor and a lower resolution with increasing distance~\cite{droeschel2014_icra}. 
In each grid cell, individual measurements are stored along with an occupancy probability and a surface element (surfel). A surfel summarizes its attributed points by their sample mean and covariance.

If available, we incorporate information from other sensors, such as an inertial measurement unit (IMU) or wheel odometry, to account for motion of the sensor during acquisition.
Furthermore, these motion estimates are used as prior for the registration. 

To register acquired 3D scans to the so far aggregated map, we use our surfel-based registration method~\cite{droeschel2014_icra,stueckler2013_mrsmaps}.
The registered 3D scan is added to the local map, replacing older measurements. 
Similar to~\cite{Hornung_octomap} we use a beam-based inverse sensor model and ray-casting to update the occupancy of a cell. 
For every measurement in the 3D scan, we update the occupancy information of cells on the ray between the sensor origin and the endpoint with an approximated 3D Bresenham algorithm~\cite{Amanatides87afast}. 
Since local multiresolution maps consist of consecutive scans from a fixed period of time, they allow for efficient local mapping with constant memory and computation demands.

Local multiresolution maps from different view poses are aligned with each other by means of surfel-based registration and build an allocentric pose graph. 
The registration result from aligning two local maps constitutes an edge in this pose graph.
Edges are added when the pose graph is extended by a new local map and between close-by local maps---\eg when the robot revisits a known location. The later allows for loop closure and minimizes the drift 
accumulated by the local mapping.  

Local maps that are added to the pose graph are subject to our refinement method, reassessing the alignment of 3D scans when more information is available. After realigning selected 3D scans from a local map, the sensor trajectory is optimized: first for refined local maps, then for the complete pose graph. \cref{fig:mapping_architecture} shows an overview of our mapping system. Since local mapping, allocentric mapping and refinement are independent from each other, our system allows for online mapping while refining previously acquired sensor data when more information is available.

\section{Hierarchical Refinement}

We model our mapping approach in a hierarchical graph-based structure as shown in \cref{fig:graph}. The coarsest level is a pose graph, representing the allocentric 6D pose of local maps $\mathcal{M} = \left\{ m_1, \ldots, m_M \right\}$ with nodes. 
Each local map aggregates multiple consecutive 3D scans and represents the robot's vicinity at a given view pose. 

They are connected by edges $\mathcal{E}_\mathcal{P}$ imposing a spatial constraint from registering two local maps with each other by surfel-based registration. 
We denote edges $E = ((M_E, M'_E),T_E,I_E) \in \mathcal{E}_\mathcal{P}$ as spatial constraint between the local maps $M_E$  and $M'_E$ with the relative pose $T_E$ and the information matrix $I_E$, which is the inverse of the covariance matrix from registration.  

The scan poses of a local map $\mathcal{M}_j$ are modeled by vertices $\mathcal{S} = \left\{ s_1, \ldots, s_S \right\}$ in a sub-graph $\mathcal{G}_j$. They are connected by edges $\mathcal{E}_\mathcal{D}$.
Registering a scan $S_E$ to a local map $M_E$ poses a spatial constraints $E = ((S_E, M_E),T_E,I_E) \in \mathcal{E}_\mathcal{D}$ with the relative pose $T_E$ and the information matrix $I_E$. 

The 3D scans of the local maps consist of a number of so-called \text{scan lines}. A scan line is the smallest element in our optimization scheme. 
Depending on the sensor setup, a scan line consists of measurements acquired in a few milliseconds. 
For the Velodyne VLP-16 used in the experiments, a scan line is a single firing sequence (1,33\,ms). 
We assume the measurements of a scan line to be too sparse for robust registration. 
Thus, we interpolate the poses of scan line acquisitions with a continuous-time trajectory representation for each sub-graph, as described later. 

Optimization of the sub-graphs and the pose graph is efficiently solved using the g$^2$o framework by \cite{kuemmerle2011_g2o}, yielding maximum likelihood estimates of the view poses $\mathcal{S}$ and $\mathcal{M}$. 
On their local time scale, sub-graphs are independent from each other, allowing to minimize errors independent from other parts of the graph. 
Optimization results from sub-graphs are incorporated in the higher level pose graph, correcting the view poses of the local maps. 
Therefore, we define the last acquired scan node in a local sub-graph as reference node and update the pose of the map node according to it.

During operation, we iteratively refine sub-graphs in parallel, depending on the available resources.
Global optimization of the full graph is only performed when the local optimization has changed a sub-graph significantly or a loop closure constraint was added.
Similarly, if global optimization was triggered by loop closure, sub-graphs are refined when the corresponding map node changed.
To determine if optimization of a sub-graph necessitates global optimization or vice-versa, we compare the refined pose of the representative scan node $s_r$ in a sub-graph to the view pose of the corresponding map node. 
For our experiments, we choose a threshold of \SI{0.01}{\metre} in translation and 1\degree in rotation.

\subsection{Local Sub-Graph Refinement}\label{sec:map_refinement}

After a local map has been added to the pose graph, the corresponding sub-graph $\mathcal{G}_\mathcal{M}$ 
is refined by realigning selected 3D scans with its local map. 
Realigning only selected 3D scans, instead of all scans in a sub-graph, allows for
fast convergence while resulting in similar map quality, as shown later in the experiments.

For a sub-graph, we determine a 3D scan $s_k$ for refinement by the spatial constraints and their associated information matrix, which is the inverse of the covariance matrix $\Sigma$ of the registration result. 
Following \cite{kerl2013dense}, we determine a scalar value for the uncertainty in the scan poses based on the entropy $H(T,\Sigma) \propto ln \left(\begin{vmatrix} \Sigma \end{vmatrix}\right)$.
It allows to select the 3D scan with the largest expected alignment error.

Furthermore, the same measure is used to compare the spatial constraints that have been added to the local map after $s_k$, to determine if realigning $s_k$ can decrease the alignment error.
The selected 3D scan is then refined, by realigning it to its local map, resulting in a refined spatial constraint in the sub-graph. 
From the sub-graph of spatial constraints, we infer the probability of the trajectory estimate given all relative pose observations
\begin{equation}
 p( \mathcal{G}_\mathcal{M} \mid \mathcal{E}_\mathcal{D} ) \propto \prod_{e_{d_{ij}} \in \mathcal{E}_\mathcal{D}} p( s_i^j \mid s_i, s_j ).
\end{equation}
We optimize the sensor trajectory for each local sub-graph independently. Results from sub-graph optimization are later incorporated when optimizing the allocentric pose graph.

\subsection{Local Window Alignment}

Registration errors are often originated from missing information in the map, \eg due to occlusions or unknown parts of the environment. Thus, registration quality can only increase if the map has been extended with measurements that provide previously unknown information. In other words, realigning a 3D scans can only increase the map quality if more scans---in best case from different view poses---have been added to the map. Therefore, we increase the local optimization window by adding 3D scans to a local map from neighboring map nodes in the higher level. For example, when the robot revisits a known part of the environment, loop closure is performed and scan nodes from neighboring map nodes are added to a local map.

\begin{figure}
{
\includegraphics[width=0.49\textwidth, trim=0cm 0.5em 0cm 0em, clip]{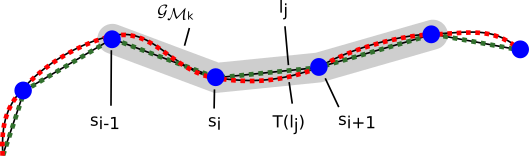}
}\\
\vspace{1cm}
\caption{Scan line poses (green squares) originated from odometry measures are refined (red squares) by interpolating with a continuous-time trajectory representation built from scan poses (blue dots) in a local sub-graph (gray). }
\label{fig:spline}
\end{figure}

\begin{figure*}[t]
  \centering
  \includegraphics[height=0.71\columnwidth]{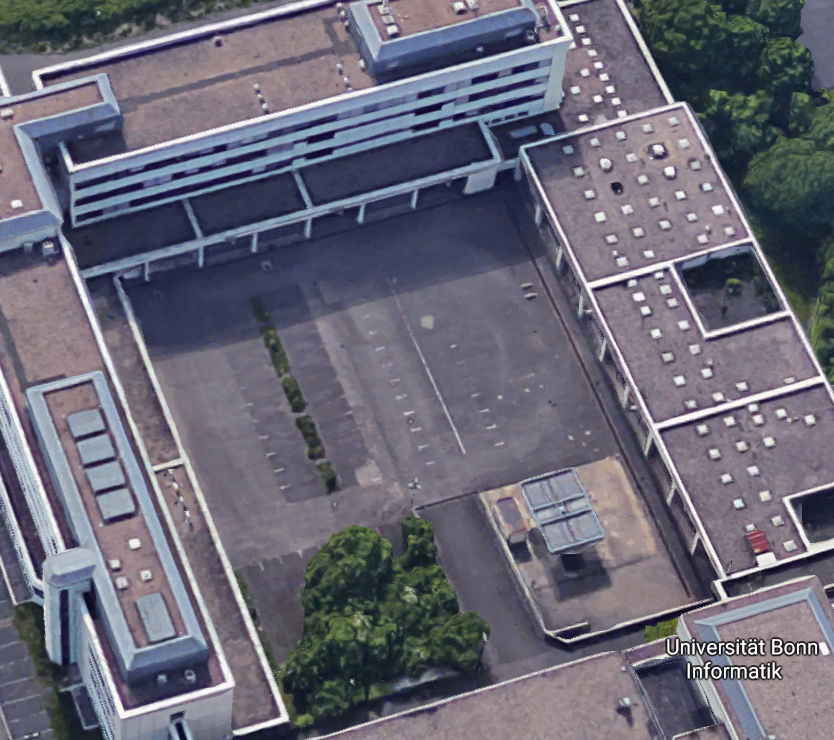}\hspace{2ex}
  \fboxsep0mm 
{
  \fbox{  \includegraphics[height=0.7\columnwidth,trim=0cm 0cm 0cm 0cm, clip]{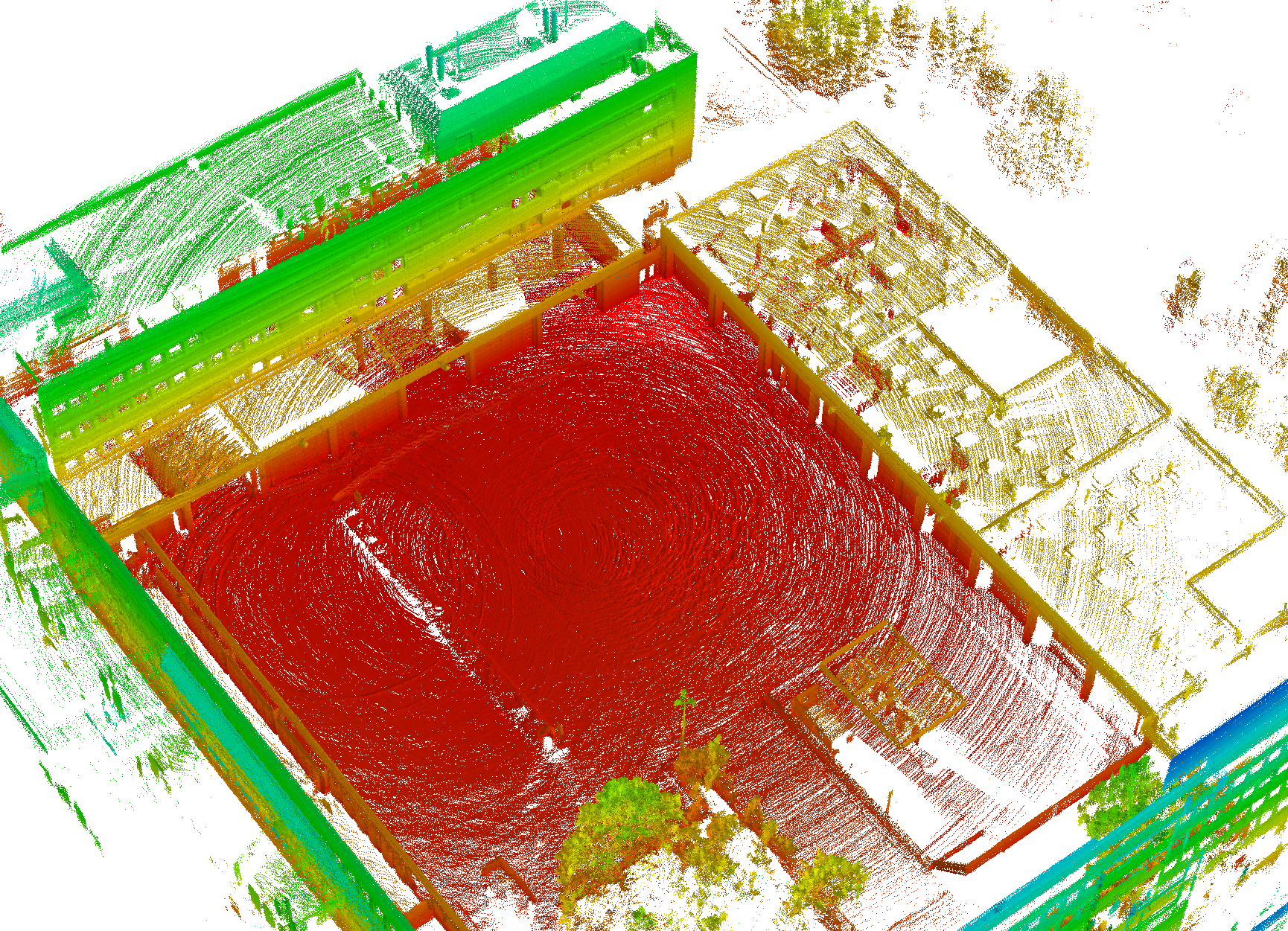}}
  }
  \caption{The resulting 3D map from an out/in-door environment. Color encodes height from the ground.}
  \label{fig:laser_map_courtyard} 
\end{figure*}

\subsection{Continuous-Time Trajectory Representation}\label{sec:continuous_trajectory}

Acquiring 3D laser scans often involves mechanical actuation, such as rotating a mirror or a diode/receiver array, during acquisition of the scan. Especially for 3D laser scans---where the acquisition of measurements for a full scan can take multiple hundred milliseconds or seconds---a discretization of the sensor pose to the time where the scan was acquired, leads to artifacts degrading the map quality. 
However, since a finer discretization of the scan poses makes the state size intractable, temporal basis functions have been used to represent the sensor trajectory \cite{Furgale6225005}. 

We represent the trajectory of the sensor as cubic B-spline in $SE(3)$ due to their smoothness and the local support property. 
The local support property allows to interpolate the trajectory from the discrete scan nodes in our local sub-graph. 
Following \cite{lovegrove2013spline}, we parameterize a trajectory by cumulative basis functions using the Lie algebra.

To estimate the trajectory spline, we use the scan nodes $s_0, \ldots, s_m$ with the acquisition times $t_{s_0}, \ldots, t_{s_m}$ as control points for the trajectory spline and denote the pose of a scan node $s_i$ as $T_{s_i}$. 
In our system, scan poses follow a uniform temporal distribution. In other words, the difference between the acquisition times of consecutive scans can be assumed to be constant. 

As illustrated in \cref{fig:spline}, we use 4 control points to interpolate the sensor trajectory between two scan nodes $s_i$ and $s_{i+1}$. 
For time $t \in \left[ t_{s_i}, t_{s_{i+1}} \right)$ the pose along the spline is defined as
\begin{equation}
T(u(t)) = T_{s_{i-1}} \prod_{j=1}^3 exp  \big( \widetilde{B}_{j}(u(t)) \Omega_{i+j}  \big).
\end{equation}
Here, $\widetilde{B}$ is the cumulative basis, $\Omega$ is the logarithmic map, and $u(t) \in \left[ 0, 1 \right)$ transforms time $t$ in a uniform time~\cite{lovegrove2013spline}. 
Finally, the spline trajectory is used to update the scan line poses between two scan nodes.

\subsection{Loop-Closure and Global Optimization}

After a new local map has been added to pose graph, we check for one new constraint between the current reference~$m_{\text{ref}}$ and other map nodes~$m_{\text{cmp}}$.
We determine a probability
\begin{equation}
  p_\text{chk}(v_{\text{cmp}}) = \mathcal{N}\left(d(m_{\text{ref}}, m_{\text{cmp}}); 0, \sigma_d^2\right)
\end{equation}
that depends on the linear distance~$d(m_{\text{ref}}, m_{\text{cmp}})$ between the view poses~$m_{\text{ref}}$ and~$m_{\text{cmp}}$.
According to~$p_\text{chk}(m)$, we choose a map node~$m$ from the graph and determine a spatial constraint between the nodes.

When a new spatial constraint has been added, the pose graph is optimized globally on the highest level.
When the optimization modifies the estimate of a map node, the changes are propagated to the sub-graph. 
Similarly, when a sub-graph changed significantly, global optimization of the highest level is carried out. 

\section{Experiments}

We assess the accuracy of our refinement method on two different data sets with different sensor setups. The first data has been recorded with a MAV equipped with a Velodyne VLP-16 lidar sensor. The second data set has been recorded in the Deutsches Museum in Munich and is provided by the Google Cartographer team~\cite{7487258}. 
Throughout the experiments, we use a distance threshold of 5\,m for adding new map nodes to the graph. 

To measure map quality, we calculate the \textit{mean map entropy} (MME)~\cite{droeschel2014_icra} from the points $\mathcal{Q} = \left\{ q_1, \ldots, q_Q \right\}$ of the resulting map. The entropy $h$ for a map point $q_k$ is calculated by
\begin{equation}
\label{eq:entropy}
h(q_k) =  \frac{1}{2} \ln\ | 2\pi e \, \Sigma(q_k)|,
\end{equation}
where $\Sigma(q_k)$ is the sample covariance of mapped points in a local radius $r$ around $q_k$.
We select $r=0.5$\,m in our evaluation.
The mean map entropy $H(\mathcal{Q})$ is averaged over all points of the resulting map
\begin{equation}
\label{eq:mapentropy}
H(\mathcal{Q}) = \frac{1}{Q} \sum_{k=1}^Q h( q_k ).
\end{equation}

It represents the \textit{crispness} or \textit{sharpness} of a map. Lower entropy measures correspond to higher map quality. 

To examine the improvement of the map quality and the convergence behavior of our method, we first run the experiments without online refinement and perform the proposed refinement as a post-processing step.
In each iteration, we refine one scan in every sub-graph and run local graph optimization. 
Local sub-graphs are refined in parallel and after refining all sub-graphs, global optimization is performed. 
To assess the number of iterations necessary for refinement, entropy measurements are plotted against the number of iterations. 
Afterwards, we run the proposed method with online refinement and compare the resulting map quality. 

Evaluation was carried out on an Intel\textsuperscript{\textregistered} Core\textsuperscript{\texttrademark} i7-6700HQ quadcore CPU running at \SI{2.6}{\giga\hertz} and \SI{32}{\giga\byte} of RAM.
For the reported runtime, we average over 10 runs for each data set.

\subsection{Courtyard}

The first data set has been recorded by a MAV during flight in a building courtyard. 
The MAV in this experiment is a DJI Matrice~600, equipped with a Velodyne~VLP-16 lidar sensor and an IMU, measuring the attitude of the robot.
The Velodyne lidar measures $\approx$~300,000 range measurements per second in 16 horizontal scan rings, has a vertical field of view of $\SI{30}{\degree}$ and a maximum range of \SI{100}{\meter}. 

It measures the environment with 16 emitter/detector pairs mounted on an array at different elevation angles from the horizontal plane of the sensor. The array is continuously rotated with up to 1200~rpm.
In our experiments, a scan line corresponds to one data packet received by the sensor, \ie 24 so-called firing sequences.
During one firing sequence (1,33\,ms) all 16 emitter/detector pairs are processed. 

In total, 2000~scans were recorded during \SI{200}{\second} flight time. 
Controlled by a human operator, the MAV traversed a building front in different heights.
The resulting graph consists of 16 map nodes with several loop closures, resulting in 27 edges between map nodes. 

\begin{figure}
 \begin{center}
  {\resizebox{0.44\textwidth}{!}{%
  \begin{tikzpicture}
    \begin{axis}[
        xlabel=Iterations,
        ylabel=Entropy]
	
    \addplot plot coordinates {
(	0	,	-0.57199593	)
(	1	,	-0.57202234	)
(	2	,	-0.59047312	)
(	3	,	-0.59329465	)
(	4	,	-0.5720249	)
(	5	,	-0.59943768	)
(	6	,	-0.5946793	)
(	7	,	-0.59500184	)
(	8	,	-0.6185839	)
(	9	,	-0.63939049	)
(	10	,	-0.64099737	)
(	11	,	-0.7395312	)
(	12	,	-0.69296055	)
(	13	,	-0.7850341	)
(	14	,	-0.79909417	)
(	15	,	-0.80288073	)
(	16	,	-0.86279222	)
(	17	,	-0.89450222	)
(	18	,	-0.95103277	)
(	19	,	-0.99550082	)
(	20	,	-0.99807178	)
(	21	,	-1.02217949	)
(	22	,	-1.00040896	)
(	23	,	-0.99359378	)
(	24	,	-0.9890628	)
(	25	,	-1.01860956	)
(	26	,	-1.01252739	)
(	27	,	-1.01450818	)
(	28	,	-1.01812392	)
(	29	,	-1.01779238	)
(	30	,	-1.02454996	)
(	31	,	-1.02014024	)
(	32	,	-1.04519768	)
(	33	,	-1.02541431	)
(	34	,	-1.02286974	)
(	35	,	-1.00860465	)
(	36	,	-1.00968651	)
(	37	,	-1.02582541	)
(	38	,	-1.01421025	)
(	39	,	-1.01704485	)
(	40	,	-1.01217099	)
(	41	,	-1.02322626	)
(	42	,	-1.02431282	)
(	43	,	-1.01351262	)
(	44	,	-1.02290981	)
(	45	,	-1.03243812	)
(	46	,	-1.01565974	)
(	47	,	-1.02759394	)
(	48	,	-1.02517209	)
(	49	,	-1.02493453	)
(	50	,	-1.01635463	)

    };
    \addlegendentry{w/o CT}
    \addplot plot coordinates {
(	0	,	-0.57199593	)
(	1	,	-0.58362326	)
(	2	,	-0.58630674	)
(	3	,	-0.61643388	)
(	4	,	-0.58471165	)
(	5	,	-0.6291602	)
(	6	,	-0.6427046	)
(	7	,	-0.65315099	)
(	8	,	-0.69924117	)
(	9	,	-0.70792247	)
(	10	,	-0.75974761	)
(	11	,	-0.76179715	)
(	12	,	-0.76012437	)
(	13	,	-0.79169061	)
(	14	,	-0.8636856	)
(	15	,	-0.88672413	)
(	16	,	-0.87999257	)
(	17	,	-0.92114473	)
(	18	,	-1.13278472	)
(	19	,	-1.17122541	)
(	20	,	-1.18542111	)
(	21	,	-1.18557845	)
(	22	,	-1.19165845	)
(	23	,	-1.20073139	)
(	24	,	-1.219611521	)
(	25	,	-1.22869978	)
(	26	,	-1.23715157	)
(	27	,	-1.23155026	)
(	28	,	-1.2371721	)
(	29	,	-1.2379849	)
(	30	,	-1.23759145	)
(	31	,	-1.23707876	)
(	32	,	-1.23794372	)
(	33	,	-1.23132574	)
(	34	,	-1.2356473	)
(	35	,	-1.2384568	)
(	36	,	-1.23685401	)
(	37	,	-1.23954508	)
(	38	,	-1.22699005	)
(	39	,	-1.23976722	)
(	40	,	-1.22700419	)
(	41	,	-1.23064604	)
(	42	,	-1.24119404	)
(	43	,	-1.2437423	)
(	44	,	-1.24022191	)
(	45	,	-1.22143941	)
(	46	,	-1.20274707	)
(	47	,	-1.21159188	)
(	48	,	-1.222272	)
(	49	,	-1.23676019	)
(	50	,	-1.24302765	)

    };
    \addlegendentry{w CT}

    \end{axis}
    \end{tikzpicture}
  }}
 \end{center}
 \caption{The resulting map entropy with and without continuous-time trajectory interpolation (CT) \cref{sec:continuous_trajectory}.}
 \label{fig:entropy_courtyard}
\end{figure}
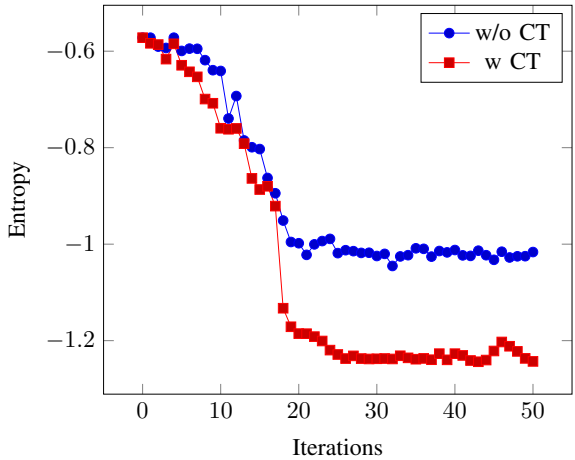

\begin{figure*}[t]
\begin{tikzpicture}
 \node[anchor=south west, inner sep=0pt] (image_map) at (0,0.5)
     {\includegraphics[width=.35\textwidth,trim=0cm 0cm 0cm 0cm, clip]{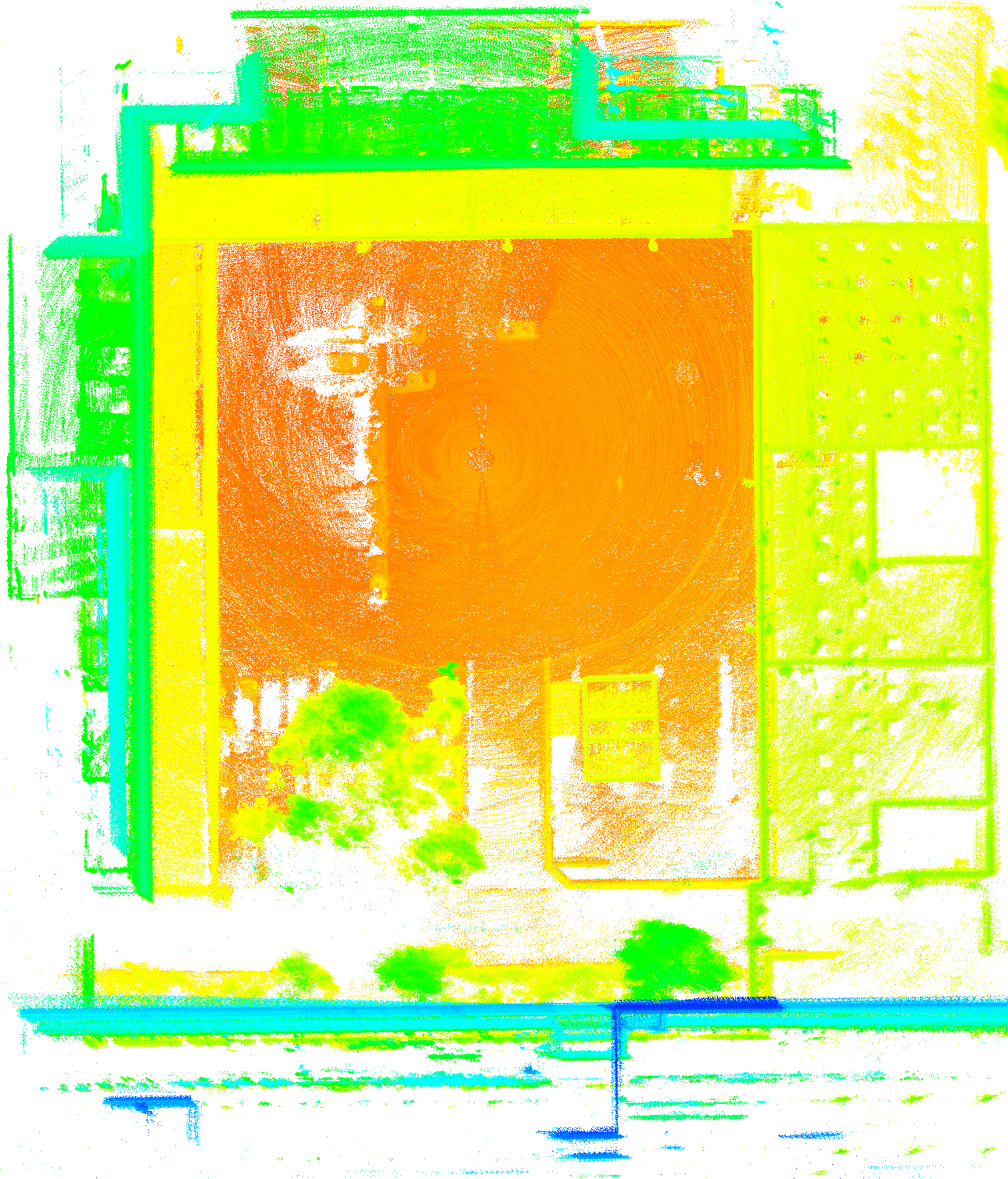}};

 \begin{scope}[
            x={(image_map.south east)},
            y={(image_map.north west)},
            font=\footnotesize\sffamily,
            line/.style={black, line width=1pt},
            every node/.style={align=center},
            box/.style={circle,minimum width=1.2cm,densely dashed,draw=black,inner sep=0.5em,very thick},
            label/.style={rectangle,rounded corners,draw=black,fill=yellow!20,text=black},
            detail_line/.style={very thick},
          ]
 	\node[box,anchor=north west] (car_detail) at (0.35, 0.74) {} ;
 	\node[box,anchor=north west] (wall_detail) at (0.68, 0.22) {} ;
 	
 	\node[inner sep=1pt, draw=black,very thick] (car) at (1.15,0.75){\includegraphics[width=.125\textwidth]{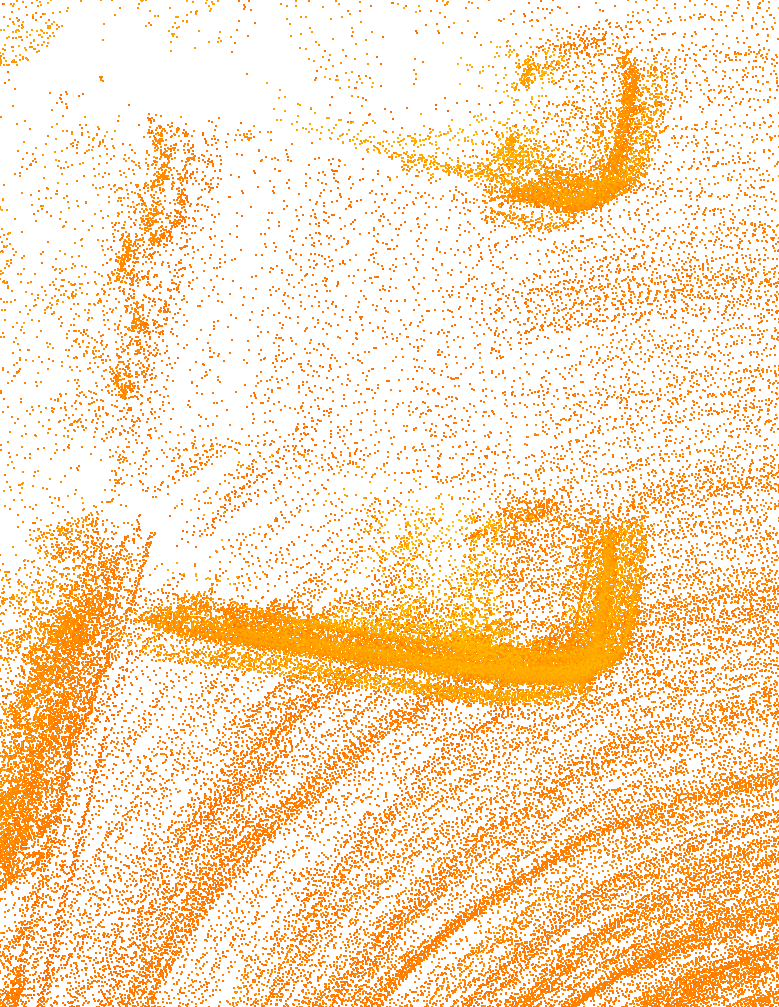}};
 	\node[inner sep=1pt, draw=black,very thick] (wall) at (1.15,0.25){\includegraphics[width=.125\textwidth]{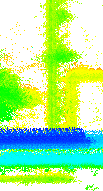}};
	\draw[detail_line] (car_detail) -- (car.west) node[midway,label] {w/o Refinement};
	\draw[detail_line] (wall_detail) -- (wall.west) ;
 \end{scope}
\end{tikzpicture}
~
\begin{tikzpicture}
 \node[anchor=south west, inner sep=0pt] (image_map) at (0,0.5)
     {\includegraphics[width=.35\textwidth,trim=0cm 0cm 0cm 0cm, clip]{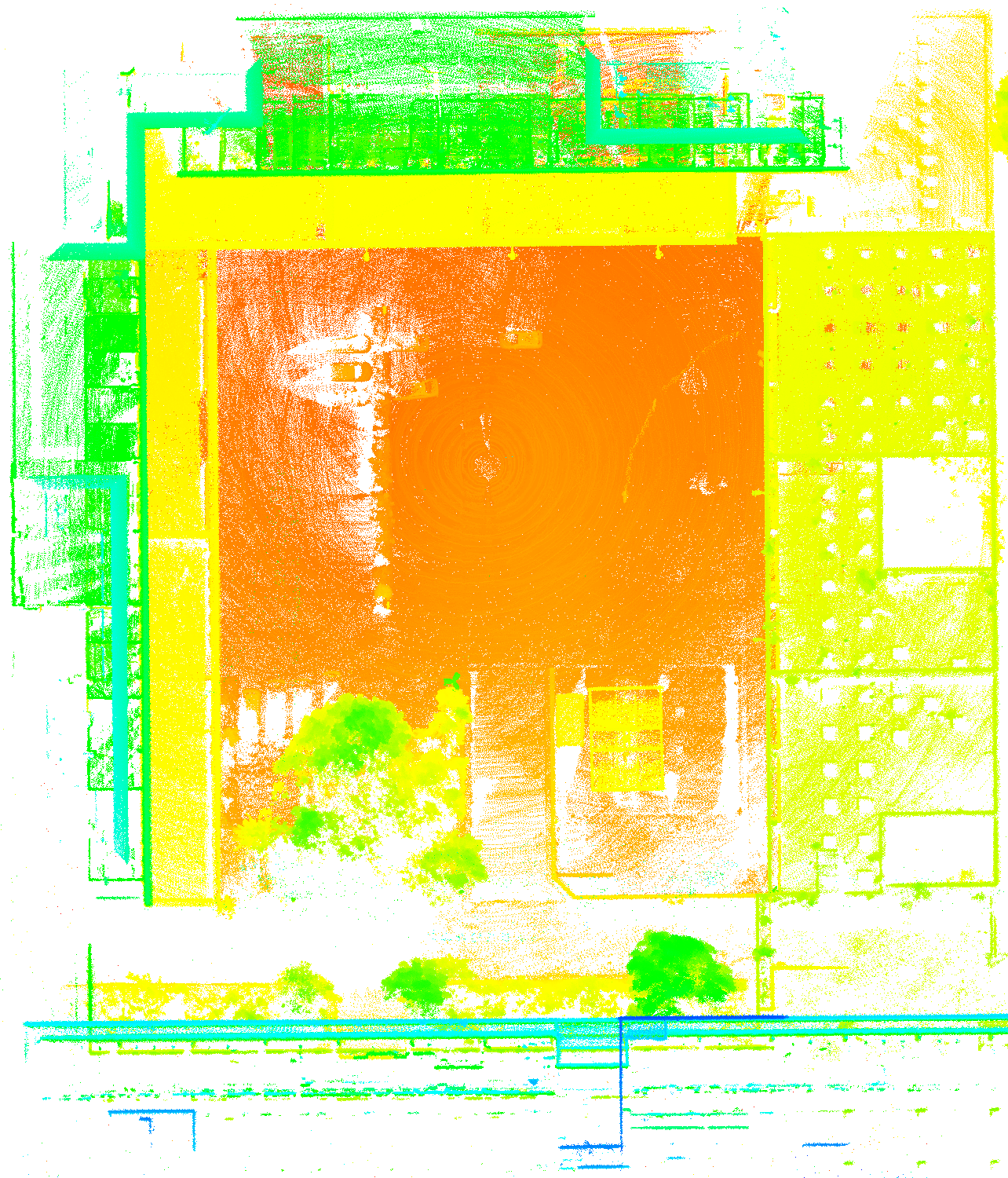}};
 \begin{scope}[
           x={(image_map.south east)},
            y={(image_map.north west)},
            font=\footnotesize\sffamily,
            line/.style={black, line width=1pt},
            every node/.style={align=center},
            box/.style={circle,minimum width=1.2cm,densely dashed,draw=black,inner sep=0.5em,very thick},
            label/.style={rectangle,rounded corners,draw=black,fill=yellow!20,text=black},
            detail_line/.style={very thick},
          ]
 	\node[box,anchor=north west] (car_detail) at (0.35, 0.74) {} ;
 	\node[box,anchor=north west] (wall_detail) at (0.68, 0.22) {} ;
 	
 	\node[inner sep=1pt, draw=black,very thick] (car) at (-0.14,0.75){\includegraphics[width=.125\textwidth]{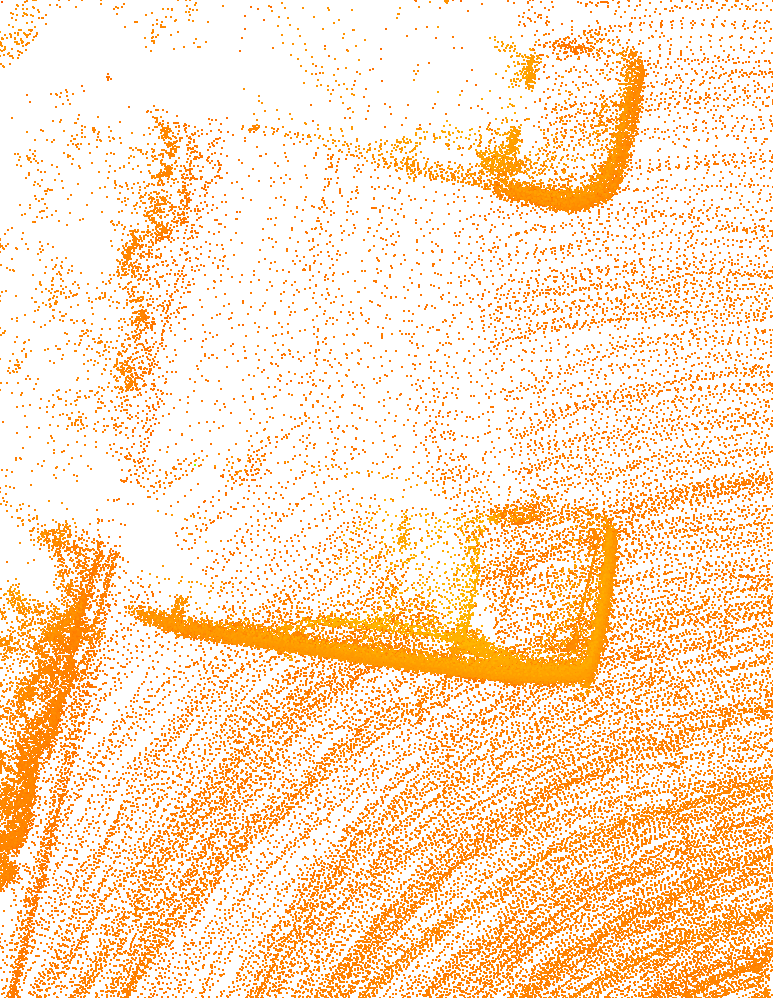}};
 	\node[inner sep=1pt, draw=black,very thick] (wall) at (-0.14,0.25){\includegraphics[width=.125\textwidth]{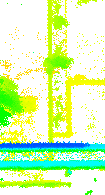}};
 	\draw[detail_line] (car_detail) -- (car.east) node[midway,label] {Refined};
	\draw[detail_line] (wall_detail) -- (wall.east) ;
 \end{scope}
\end{tikzpicture}
\caption{Resulting point clouds from the Courtyard data set. Left: results from our previous method~\cite{Droeschel2017104}. Right: results from the presented method. Color encodes the height. }
  \label{fig:refine_courtyard}
\end{figure*}
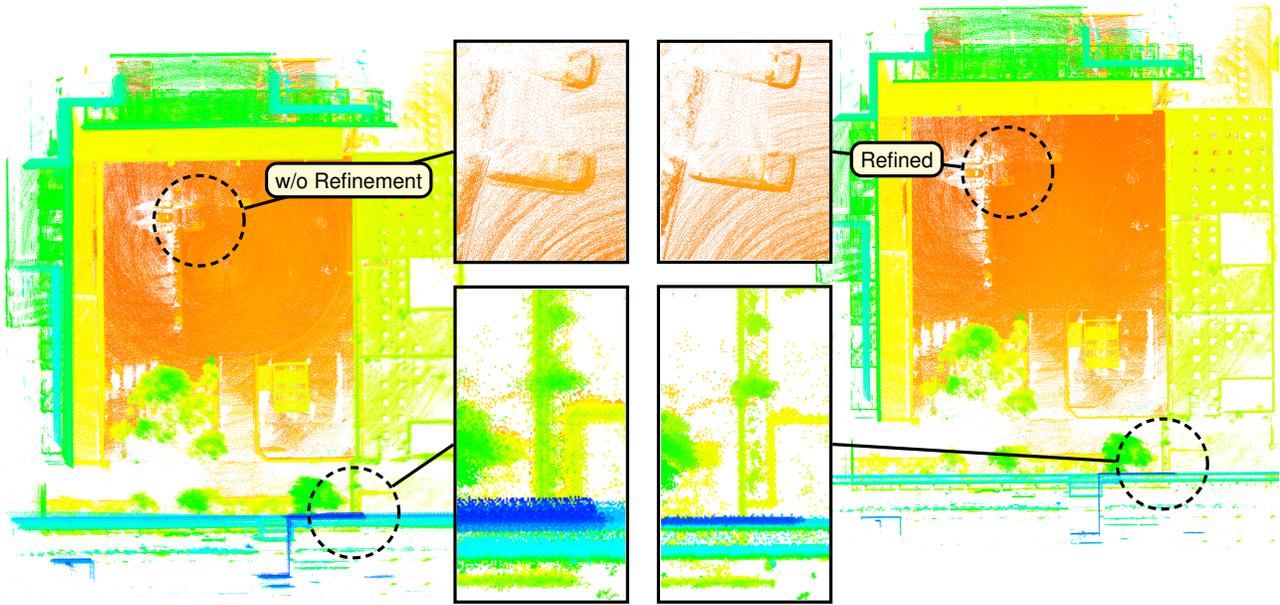

\cref{fig:laser_map_courtyard} shows the environment and a resulting map. 
In a first experiment, we compare the method from our previous work with the proposed method. 
The resulting point clouds are shown in \cref{fig:refine_courtyard}. 
The figure shows, that the proposed method corrects misaligned 3D scans and increases the map quality. 
\cref{fig:entropy_courtyard} shows the resulting entropy plotted against the number of iterations when running the refinement as post-processing step. 
During one iteration, a single 3D scan in each map node is refined. 
We measure an average runtime of \SI{54}{\milli\second} per iteration for refining a single map node and \SI{380}{\milli\second} per iteration for refining all 16 map nodes in parallel.

\subsection{Deutsches Museum}

For further evaluation, we compare our method on a data set that has been recorded at the Deutsches Museum in Munich. The data set is provided by Hess~\etal~\cite{7487258}. Two Velodyne VLP-16 mounted on a backpack are carried through the museum. Parts of the data set contain dynamic objects, such as moving persons. The provided data set includes a calibration between the two laser scanners---one mounted horizontal, one vertical. We use the provided calibration between the two sensors as initial calibration guess and refine it by adding additional constraints to our pose graph and the local sub-graphs. Similar to the alignment of two local maps, we refine the calibration parameters by our surfel-based method, registering the scans of the graph from the horizontal scanner to the scans of the graph from the vertical scanner. 

\begin{figure}
 \begin{center}
   {\resizebox{0.44\textwidth}{!}{%
  \begin{tikzpicture}
    \begin{axis}[
        xlabel=Iterations,
        ylabel=Entropy]
    \addplot plot coordinates {

(0, -2.12754227)
(1, -2.17798824)
(2, -2.19492671)
(3, -2.20238990)
(4, -2.21983976)
(5, -2.24335992)
(6, -2.25538508)
(7, -2.26518642)
(8, -2.26390758)
(9, -2.25270176)
(10, -2.24894839)
(11, -2.26320211)
(12, -2.26963542)
(13, -2.27786642)
(14, -2.27616185)
(15, -2.29267169)
(16, -2.29561492)
(17, -2.30567314)
(18, -2.30799572)
(19, -2.31565611)
(20, -2.32239921)
(21, -2.32659141)
(22, -2.31845054)
(23, -2.31606034)
(24, -2.32221314)
(25, -2.32503454)
(26, -2.31990025)
(27, -2.31824584)
(28, -2.32048339)
(29, -2.32507825)
(30, -2.31520511)
(31, -2.32668189)
(32, -2.33155374)
(33, -2.32978796)
(34, -2.32914720)
(35, -2.34244985)
(36, -2.33924588)
(37, -2.34082428)
(38, -2.33974631)
(39, -2.34190124)
(40, -2.33497753)
(41, -2.33226007)
(42, -2.33568532)
(43, -2.33283389)
(44, -2.32925736)
(45, -2.33306040)
(46, -2.33920143)
(47, -2.33939474)
(48, -2.34327439)
(49, -2.34106753)
(50, -2.34413333)
    };
    \addlegendentry{w/o COV Selection, w/o CT}

    \addplot
        plot coordinates {        
  
(0, -2.12756200)
(1, -2.18625466)
(2, -2.23908857)
(3, -2.27698450)
(4, -2.28314643)
(5, -2.28990335)
(6, -2.29529741)
(7, -2.31503746)
(8, -2.32101931)
(9, -2.33080878)
(10, -2.34095579)
(11, -2.34628515)
(12, -2.34810204)
(13, -2.34682366)
(14, -2.35193087)
(15, -2.35007169)
(16, -2.35093848)
(17, -2.35102455)
(18, -2.35127419)
(19, -2.35003950)
(20, -2.35002682)
(21, -2.35072153)
(22, -2.34980721)
(23, -2.34821293)
(24, -2.35201508)
(25, -2.35177249)
(26, -2.35068773)
(27, -2.35248687)
(28, -2.35147785)
(29, -2.34792766)
(30, -2.34917409)
(31, -2.34860007)
(32, -2.34957866)
(33, -2.34756163)
(34, -2.34714171)
(35, -2.34737089)
(36, -2.34858285)
(37, -2.34913736)
(38, -2.35028379)
(39, -2.35171805)
(40, -2.35076127)
(41, -2.34974084)
(42, -2.35673906)
(43, -2.35807820)
(44, -2.34929077)
(45, -2.35234846)
(46, -2.34791075)
(47, -2.34755434)
(48, -2.34800212)
(49, -2.35174063)
(50, -2.35010792)

        };
    \addlegendentry{w/~ COV Selection, w/o CT}

        \addplot
        plot coordinates {  
  
(0, -2.12754227)
(1,-2.18009411)
(2,-2.20779869)
(3,-2.22784163)
(4,-2.25396756)
(4,-2.28350951)
(5,-2.29467129)
(6,-2.32738513)
(7,-2.33080364)
(8,-2.33491797)
(9,-2.33553637)
(10,-2.33838508)
(11,-2.34089309)
(12,-2.34197464)
(13,-2.3477527)
(14,-2.34790994)
(15,-2.34985845)
(16,-2.35377309)
(17,-2.35385972)
(18,-2.35778862)
(19,-2.36328419)
(20,-2.37354709)
(21,-2.37760018)
(22,-2.38124825)
(23,-2.39363084)
(24,-2.39317358)
(25,-2.39572345)
(26,-2.40331286)
(27,-2.41490116)
(28,-2.41507037)
(29,-2.41788834)
(30,-2.41863604)
(31,-2.41870611)
(32,-2.41865718)
(33,-2.41978084)
(34,-2.41988356)
(35,-2.41960923)
(36,-2.4193937)
(37,-2.41959088)
(38,-2.4187386)
(39,-2.41807227)
(40,-2.41701112)
(41,-2.41979367)
(42,-2.41968893)
(43,-2.42028649)
(44,-2.41962823)
(45,-2.41971107)
(46,-2.42128249)
(47,-2.42034415)
(48,-2.41937496)
(49,-2.42051255)
(50,-2.42137072)

};
\addlegendentry{w/~ COV Selection, w/~ CT}

    \end{axis}
    \end{tikzpicture}
  }}
 \end{center}
 \caption{The resulting map entropy with and without our covariance based scan selection (COV Selection) described in \cref{sec:map_refinement} and continuous-time trajectory interpolation (CT) \cref{sec:continuous_trajectory}. Scan selection leads to faster convergence of our method.}
 \label{fig:entropy_scan_selection}
\end{figure}
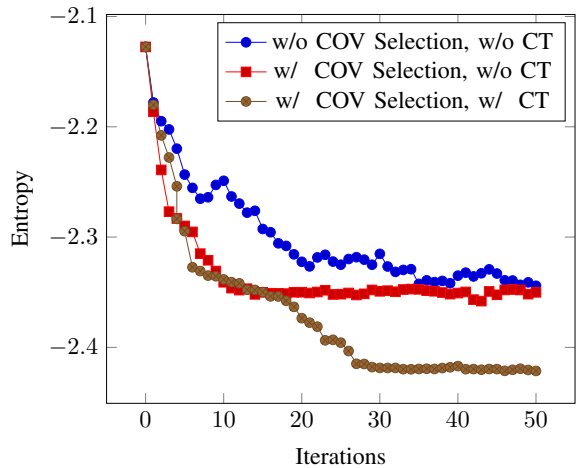

Following \cite{nuchter2017improving}, we select a part of the data set and run our method on it. Besides visual inspection of the resulting point cloud we compute the entropy as described before. 
\cref{fig:entropy_scan_selection} shows the convergence behavior of our method with and without the covariance-based scan selection. It indicates that our covariance-based scan selection leads to faster convergence. 
Furthermore, we compare the presented method with the method from our previous work.
We also compare our method to Google's Cartographer~\cite{7487258} and the continuous-time slam method from~\cite{elseberg2013algorithmic} that has been evaluated in~\cite{nuchter2017improving}. 
We summarize our results for each method in \cref{tab:entropy}.

\begin{table}[t]
  \centering
  \caption{Resulting best mean map entropies (MME) for the data set recorded at Deutsches Museum.}
  \begin{tabular}{lcc} %
    \toprule
    \textbf{Method} & & \textbf{MME}\\
    \midrule
    Cartographer~\cite{7487258} & & -2.04 \\ 
    Droeschel~\etal~\cite{Droeschel2017104} & & -2.12 \\ 
    N\"uchter~\etal~\cite{nuchter2017improving}& & -2.34 \\ 
    Ours & & -2.42 \\ 
    \bottomrule
  \end{tabular}
  \label{tab:entropy}
\end{table}

\begin{figure*}[t]
\begin{tikzpicture}
 \node[anchor=south west, inner sep=0pt] (image_map) at (0,0.5)
     {\includegraphics[width=.95\textwidth,trim=0cm 0cm 0cm 0cm, clip]{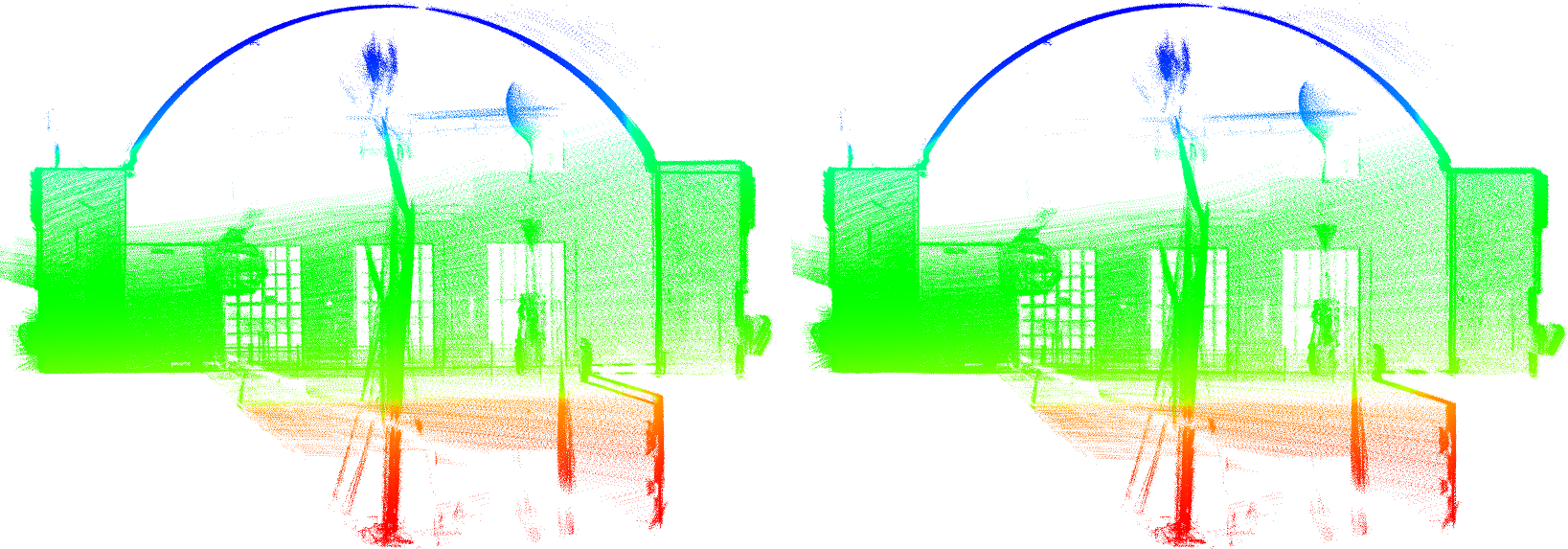}};

 \begin{scope}[
            x={(image_map.south east)},
            y={(image_map.north west)},
            font=\footnotesize\sffamily,
            line/.style={red, line width=1pt},
            every node/.style={align=center},
            box/.style={circle,minimum width=1.2cm,densely dashed,draw=red,inner sep=0.5em,very thick},
            label/.style={rectangle,rounded corners,draw=black,fill=yellow!20,text=black},
            detail_line/.style={very thick},
          ]
 	\node[box,anchor=north west] (nuechter_detail_1) at (0.425, 0.74) {} ;
 	\node[box,anchor=north west] (nuechter_detail_2) at (0.38, 0.37) {} ;
 \end{scope}
\end{tikzpicture}
\begin{tikzpicture}
 \node[anchor=south west, inner sep=0pt] (image_map) at (0,0.5)
     {\includegraphics[width=.95\textwidth,trim=0cm 0cm 0cm 0cm, clip]{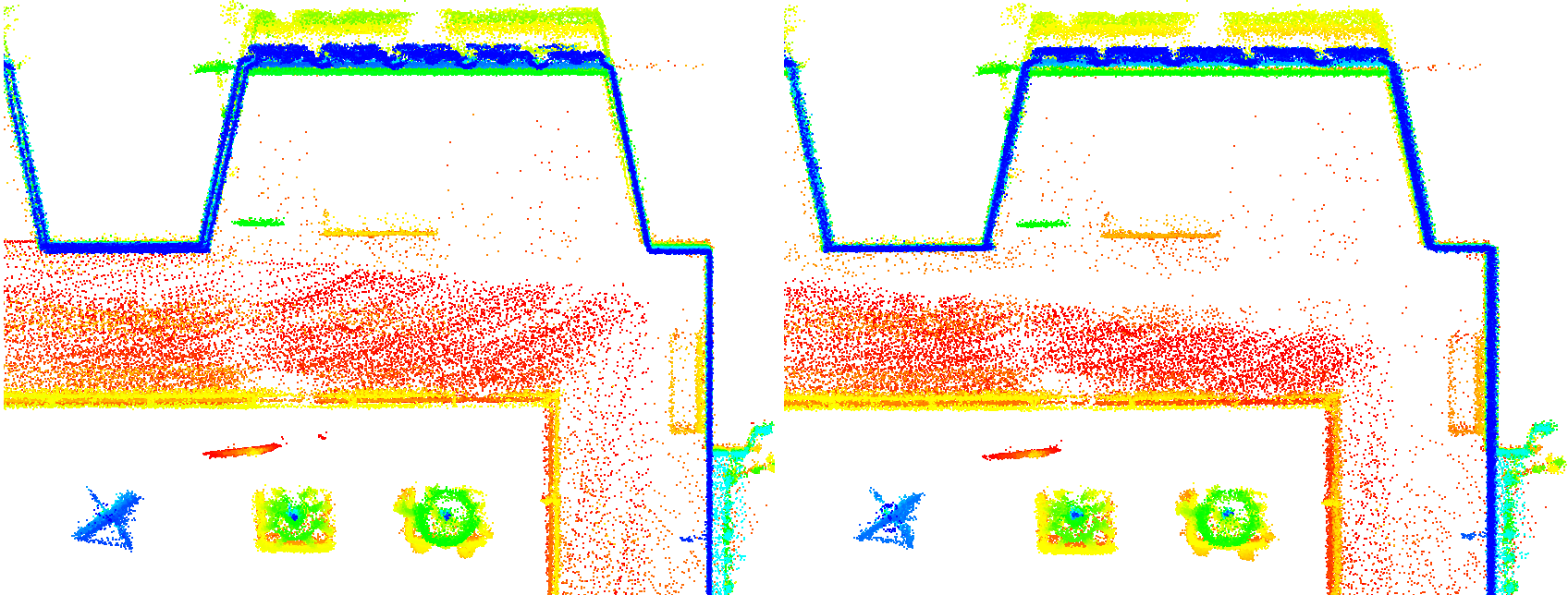}};
 \begin{scope}[
            x={(image_map.south east)},
            y={(image_map.north west)},
            font=\footnotesize\sffamily,
            line/.style={red, line width=1pt},
            every node/.style={align=center},
            box/.style={circle,minimum width=0.7cm,densely dashed,draw=red,inner sep=0.5em,very thick},
            label/.style={rectangle,rounded corners,draw=black,fill=yellow!20,text=black},
            detail_line/.style={very thick},
          ]
 	\node[box,anchor=north west] (scoop_detail) at (0.425, 0.62) {} ;
 \end{scope}
\end{tikzpicture}
\caption{Resulting point clouds from a part of the trajectory of the Deutsches Museum data set. Left: results from \cite{nuchter2017improving}. Right: results from our method. Red (dashed) circles highlight distorted parts of the map. Color encodes the height. }
\end{figure*}

\section{Conclusions}

In this paper, we proposed a hierarchical, continuous-time approach for laser-based 3D SLAM. 
Our method is based on efficient local mapping and a hierarchical optimization back-end.
Measurements of a 3D laser scanner are aggregated in local multiresolution maps, by means of surfel based registration. The local maps 
are used in a graph-based structure for allocentric mapping. 
The individual 3D scans in the local map model a sub-graph to 
incorporate corrections when refining these sub-graphs. 
Graph optimization is performed to account for drift and misalignments in the local maps.
Furthermore, a continuous-time trajectory representation allows to interpolate measurements between discrete scan poses.
Evaluation shows that our approach increases map quality and leads to sharper maps.

                                  %
                                  %
                                  %
                                  %
                                  %

%

%

%

\section*{Acknowledgments}
This work was supported by grant BE 2556/7 (Mapping on Demand) of the German Research Foundation (DFG), grant 01MA13006D (InventAIRy) of the German BMWi, and grant 644839 (CENTAURO) of the European Union's Horizon 2020 Programme.
We like to thank Andreas N{\"u}chter, Michael Bleier and Andreas Schauer for providing datasets for the evaluation.

\bibliographystyle{IEEEtran}
\bibliography{references}

\end{document}